\documentclass[letterpaper, 10 pt, conference]{ieeeconf}
\IEEEoverridecommandlockouts
\usepackage[usenames]{xcolor}
\usepackage{amsmath}
\usepackage{amssymb}
\usepackage{diagbox}
\usepackage{epsfig}
\usepackage{graphicx}
\usepackage{mathptmx}
\usepackage{times}
\usepackage{url}
\usepackage{multirow, makecell}
\usepackage{subcaption}
\usepackage{hyperref}
\usepackage{tabularray}
\usepackage[normalem]{ulem} 
\usepackage{xspace}
\usepackage[switch]{lineno}
\usepackage[final]{changes} 
\usepackage{cuted}
\usepackage[flushleft]{threeparttable}

\usepackage{xifthen}
\usepackage{xparse}
\usepackage{subdepth}
\usepackage{leftidx}
\usepackage{bm}

\newcommand{\bbm}{\begin{bmatrix}}
\newcommand{\ebm}{\end{bmatrix}}

\DeclareMathAlphabet{\mbf}{OT1}{ptm}{b}{n}
\newcommand{\mbs}[1]{{\bm{#1}}} 

\newcommand{\mbsbar}[1]{{\overline{\boldsymbol{#1}}}}
\newcommand{\mbshat}[1]{{\hat{\boldsymbol{#1}}}}
\newcommand{\mbstilde}[1]{{\tilde{\boldsymbol{#1}}}}

\newcommand{\mbsdot}[1]{{\dot {\boldsymbol{#1}}}}

\newcommand{\mbfbar}[1]{{\overline{\mbf{#1}}}}
\newcommand{\mbfhat}[1]{{\hat{\mbf{#1}}}}
\newcommand{\mbftilde}[1]{{\tilde{\mbf{#1}}}}

\newcommand{\mbfdot}[1]{{\dot{\mbf{#1}}}}

\newcommand{\cframe}[1]{{\smash{\protect\underrightarrow{\mathcal{F}}_{#1}}}}

\DeclareMathAlphabet{\mathbfit}{OML}{cmm}{b}{it}
\newcommand{\homo}[1]{{\mathbfit{#1}}}

\newcommand{\mbfh}[1]{{\homo{#1}}}

\newcommand{\pos}[2]{\leftidx{_{#1}}{ \mbf r}{_{#2}}} 
\newcommand{\posh}[2]{\leftidx{_{#1}}{\mbfh r}{_{#2}}} 

\newcommand{\vel}[3]{\leftidx{_{#1}}{\mbf v}{\IfValueTF{#2}{_{#2#3\hspace{2pt}}}{}}} 
\newcommand{\velbar}[3]{\leftidx{_{#1}}{\mbfbar v}{\IfValueTF{#2}{_{#2#3\hspace{2pt}}}{}}} 
\newcommand{\velhat}[3]{\leftidx{_{#1}}{\mbfhat v}{\IfValueTF{#2}{_{#2#3\hspace{2pt}}}{}}} 
\newcommand{\veldot}[3]{\leftidx{_{#1}}{\mbfdot v}{\IfValueTF{#2}{_{#2#3\hspace{2pt}}}{}}} 

\newcommand{\acc}[3]{\leftidx{_{#1}}{\mbf a}{\IfValueTF{#2}{_{#2#3\hspace{2pt}}}{}}} 
\newcommand{\acctilde}[3]{\leftidx{_{#1}}{\mbftilde a}{\IfValueTF{#2}{_{#2#3\hspace{2pt}}}{}}} 
\newcommand{\accbar}[3]{\leftidx{_{#1}}{\mbfbar a}{\IfValueTF{#2}{_{#2#3\hspace{2pt}}}{}}} 

\newcommand{\rotvel}[3]{\leftidx{_{#1}}{\mbs \omega}{\IfValueTF{#2}{_{#2#3\hspace{2pt}}}{}}} 
\newcommand{\rotveltilde}[3]{\leftidx{_{#1}}{\mbstilde \omega}{\IfValueTF{#2}{_{#2#3\hspace{2pt}}}{}}} 
\newcommand{\rotvelbar}[3]{\leftidx{_{#1}}{\mbsbar \omega}{\IfValueTF{#2}{_{#2#3\hspace{2pt}}}{}}} 
\newcommand{\rotvelhat}[3]{\leftidx{_{#1}}{\mbshat \omega}{\IfValueTF{#2}{_{#2#3\hspace{2pt}}}{}}} 
\newcommand{\rotveldot}[3]{\leftidx{_{#1}}{\mbsdot \omega}{\IfValueTF{#2}{_{#2#3\hspace{2pt}}}{}}} 

\newcommand{\T}[2]{\leftidx{}{\mbfh T}{_{#1#2\hspace{2pt}}}} 

\newcommand{\voxblox}{\emph{voxblox}\xspace}
\newcommand{\voxgraph}{\emph{voxgraph}\xspace}

\newcommand{\dimt}[1]{\textcolor{gray}{#1}}

\title{\LARGE \bf
Scalable Outdoors Autonomous Drone Flight with Visual-Inertial SLAM and Dense Submaps Built without LiDAR}

\author{Sebasti\'an Barbas Laina$^{1,4}$,
Simon Boche$^{1}$,
Sotiris Papatheodorou$^{1,2,4,5}$,\\
Dimos Tzoumanikas$^{2}$,
Simon Schaefer$^{1}$,
Hanzhi Chen$^{1}$
and Stefan Leutenegger$^{1,2,3,4,5}$
\thanks{This work was supported by the Technical University of Munich, MIRMI, the TUM Innovation Network CoConstruct, Leica Geosystems AG, and the EU Horizon project DigiForest.}
\thanks{$^{1}$Mobile Robotics Lab, School of Computation, Information and Technology, Technical University of Munich. E-mail addresses: \texttt{\{sebastian.barbas, simon.boche, sotiris.papatheodorou, simon.k.schaefer, hanzhi.chen, stefan.leutenegger\}@tum.de}}%
\thanks{$^{2}$Smart Robotics Lab, Department of Computing, Imperial College London. E-mail addresses: \texttt{\{s.papatheodorou18, dimosthenis.tzoumanikas14, s.leutenegger\}@ic.ac.uk}}%
\thanks{$^{3}$Mobile Robotics Lab, Department of Mechanical and Process Engineering, ETH Zurich. E-mail addresses: \texttt{lestefan@ethz.ch}}
\thanks{$^{4}$Munich Institute of Robotics and Machine Intelligence (MIRMI).}%
\thanks{$^{5}$Munich Center for Machine Learning (MCML).}%
}

\begin{document}

\maketitle
\begin{strip}
    \vspace{-4.0cm}
\end{strip}
\thispagestyle{empty}
\pagestyle{empty}

\begin{abstract}

 Autonomous navigation is needed for several robotics applications. In this paper we present an autonomous Micro Aerial Vehicle (MAV) system which purely relies on cost-effective and light-weight passive visual and inertial sensors to perform large-scale autonomous navigation in outdoor, unstructured and cluttered environments. We leverage visual-inertial simultaneous localization and mapping (VI-SLAM) for accurate MAV state estimates and couple it with a volumetric occupancy submapping system to achieve a scalable mapping framework which can be directly used for path planning. To ensure the safety of the MAV during navigation, we also propose a novel reference trajectory anchoring scheme that deforms the reference trajectory the MAV is tracking upon state updates from the VI-SLAM system in a consistent way, even upon large state updates due to loop-closures. We thoroughly validate our system in both real and simulated forest environments and at peak velocities up to 3 m/s -- while not encountering a single collision or system failure. To the best of our knowledge, this is the first system which achieves this level of performance in such an unstructured environment using low-cost passive visual sensors and fully on-board computation, including VI-SLAM. Code available at \href{https://github.com/ethz-mrl/mrl_navigation}{https://github.com/ethz-mrl/mrl\_navigation}.

\end{abstract}
\section{Introduction}

Autonomous navigation is essential for many robot-based applications, allowing the robot to move freely and increasing the types of tasks it can perform. More specifically, MAVs have proven to be an ideal solution for search and rescue, autonomous inspection or environmental monitoring due to their superior motion capabilities, which allow them to navigate through complex, cluttered, and unstructured environments. Outdoor environments are often especially challenging due to their scale, requiring high flight speeds to traverse before the limited battery capacity is exhausted. Forests are especially demanding as they combine all aforementioned challenges while being GPS-denied.

\begin{figure}[t]
    \centering
    \includegraphics[width=\columnwidth, trim={7cm, 11cm, 1cm, 9cm}, clip]{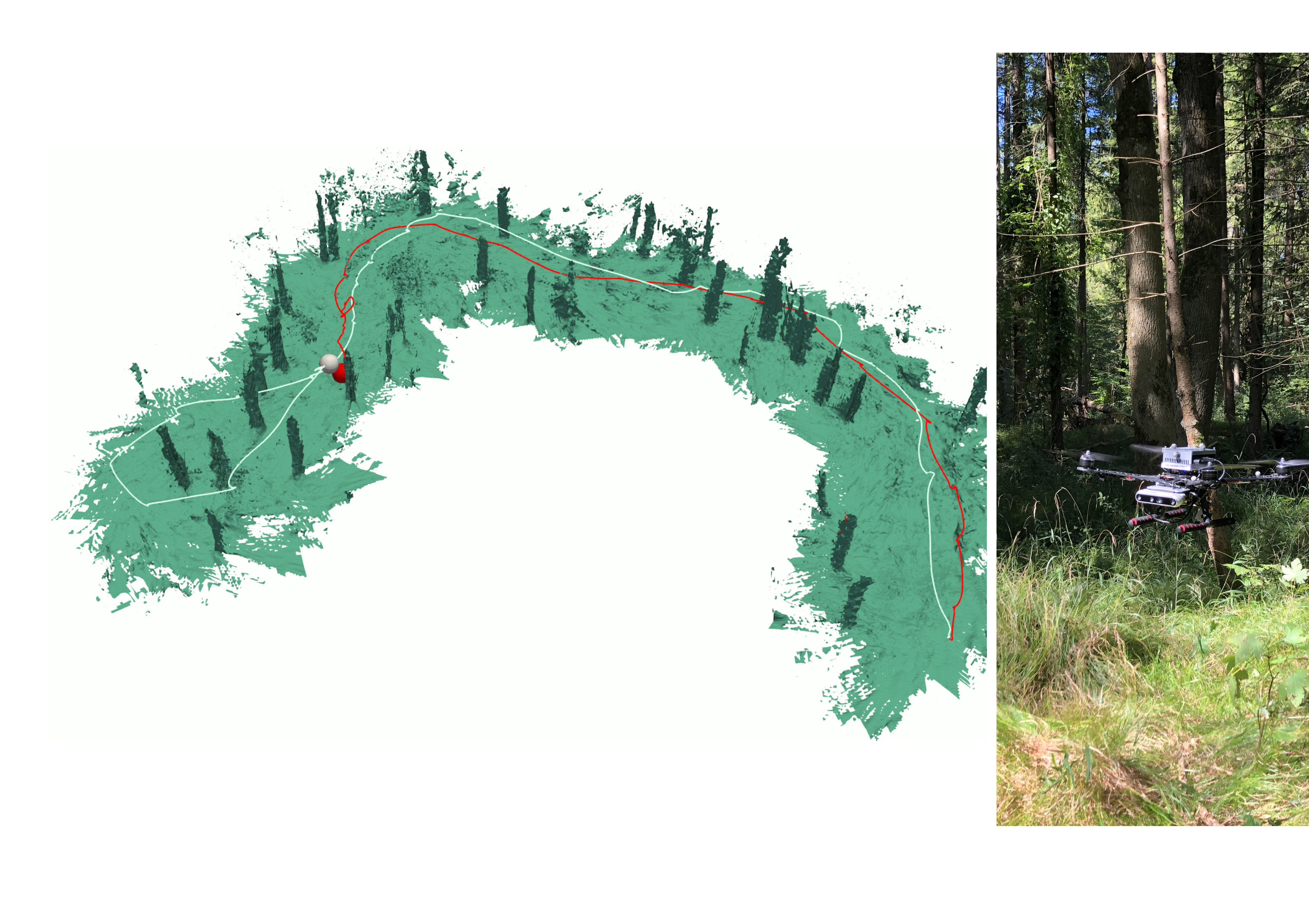}
    \caption{Mesh of a real-world environment generated using our proposed approach. The executed mission consists of a back-and-forth trajectory with a small loop at the end. The red line shows the drone trajectory flying out, and the white one is the returning trajectory. The red sphere represents the start of the mission and the white one the end of it. Right: the drone used in these real-world experiments.}
    \label{fig:mesh_reconstruction}
\end{figure}

To fly mapping missions autonomously, aerial vehicles require robust capabilities of state estimation, dense mapping, and a flight controller that can track the commanded path. At the same time odometry drifts are unavoidable and need to be corrected, especially upon loop-closure, in order to maintain a consistent map representation for effective navigation and to communicate it to end-users.

State-of-the-art autonomous drone flying in forests typically relies on LiDAR for precise and far-reaching range measurements \cite{liu2022large,tranzatto2022cerberus}. In this work, we aim to overcome such constraints by employing only passive vision and an Inertial Measurement Unit (IMU) in a quest towards cheaper, lighter, and more scalable drone systems.
In short, the main contributions of this paper are:
\begin{itemize}
  \item To the best of our knowledge, we present the \emph{first} autonomous drone system that can navigate in \added{large-scale outdoor} unstructured environments while leveraging a SLAM system with integrated loop-closure correction and only relying on passive visual and inertial sensors. It also performs learnt depth estimation, driving volumetric occupancy mapping in submaps, path planning, and trajectory tracking -- with all computations on-board.
  \item We introduce a novel approach to trajectory deformation at odometry rate to continue performing trajectory tracking upon state estimation updates, most pronounced when loop-closures occur. This allows us to track planned paths without stopping upon odometry shifts.
  \item We demonstrate our approach in simulated and real-world forests of up to 467 trees per hectare and achieve peak velocities of up to 3 m/s without collisions.
  \item We perform a mapping completeness and accuracy analysis to demonstrate the benefits of online maps generated via SLAM poses instead of visual-inertial odometry. We demonstrate that with more loop-closures, the environment reconstruction quality is improved.
\end{itemize}

\section{Related Work}
\label{section:related_work}

In this section we will present an overview of complete frameworks for autonomous navigation, regardless of the specific sensors employed. Our goal is to create a vision-based system for autonomous navigation of MAVs that can operate in large, unstructured, and cluttered outdoor environments while building a map  for downstream tasks, flying as fast as possible and with all computations on-board.

Several systems which leverage vision sensors for autonomous navigation have been proposed in the literature. One of the earliest versions of vision based navigation was presented in \cite{faessler2016autonomous}. Based on a monocular camera and IMU sensors, the system performed state-estimation for trajectory tracking while streaming data via WiFi to an off-board laptop to perform dense 3D reconstruction. This method highlighted the potential of autonomous navigation based on vision systems, although it required a reliable WiFi connection and an operator to provide trajectories for the MAV.

A system closely aligned with our work is \cite{oleynikova2020open}, where an MAV based on visual-inertial sensors performs autonomous navigation while generating a dense map for path planning. The system relies on depth data from RGB-D sensors, estimates MAV poses using visual-inertial odometry (VIO) \cite{schneider2018maplab} and the information is aggregated into a monolithic map which represents the environment as a Euclidean signed distance field (ESDF), based on \cite{voxblox} and \cite{millane2018cblox}. Similarly, Lin \textit{et al.} \cite{lin2018autonomous} present a system based on a visual-inertial sensor to build a monolithic 3D map based on OctoMap\cite{hornung2013octomap}.

Monolithic maps are suboptimal for outdoor autonomous navigation for several reasons. First is the necessity for a larger map as the mission's scale increases, leading to slower data integration. Secondly, as the odometry drift accumulates during navigation, erroneously repeated geometric structures will appear within the map. The lack of a corrective mechanism, such as loop-closures, leads to an artificially cluttered map, hindering path planning. The issue worsens upon geometrically discontinuous odometry estimates, which typically occur after SLAM loop-closures. As we will demonstrate, drift is the main source of error in map reconstruction, \added{potentially making the overall map unusable when the error accumulates over long missions. This makes drift} correcting techniques \replaced{essential}{necessary} for accurate reconstructions \added{in outdoor environments}. Nonetheless, other works \cite{Cieslewski2017, Zichao2018}, \added{\cite{ foehn2022agilicious, campos2021autonomous, alarcon2021efficient, reijgwart2024waverider, bircher2016receding, liu2024omninxt} still} rely on VIO, instead of SLAM. To circumvent these issues, we couple volumetric occupancy submaps to a SLAM system.

\added{To the best of our knowledge, the only work that addresses autonomous navigation with loop closures based on visual data is the autonomous exploration method presented by Lee \textit{et al.} \cite{real}. Nonetheless, the proposed method is only suited for indoor environments with relatively small drift, since it leverages a monolithic map and it does not employ any safety mechanisms during trajectory tracking, like our proposed trajectory deformation. At the same time, it only achieves a maximum flight speed of 0.5 m/s.} 

As a follow up to \cite{oleynikova2020open}, a submap-based, dense mapping system developed from \voxblox \cite{voxblox} was presented in \voxgraph \cite{reijgwart2019voxgraph}, which leverages the internal ESDF mapping representation to perform submap to submap alignment, reducing the mapping inconsistencies due to drift.
This system was then used in \cite{glocal} to perform exploration.
The main differences with our work are: \cite{glocal} leverages VIO as a state-estimator, instead of SLAM, for their real-world experiments, and the submap generation strategy is based on time while ours is based on the keyframes detected by the SLAM framework which is more suitable for large-scale missions as it generates fewer submaps, reducing memory usage.

Research on autonomous navigation in cluttered and unstructured environments based on visual sensors has gained interest in recent years, although none of the works focused on the problem of mapping accuracy or in leveraging SLAM. Prominent works in this area are \cite{zhou2021ego} and \cite{loquercio2021learning}. Zhou \textit{et al.} \cite{zhou2021ego} build on top of \cite{zhou2020ego} to enable swarms of MAVs to navigate the environment, although they rely on VIO, construct a monolithic map and only achieve peak velocities of 1.5 m/s, lower than our maximum 3 m/s. In \cite{loquercio2021learning} they tackle the issue of high-speed navigation by leveraging an end-to-end deep learning approach \replaced{that follows reference trajectories given by an operator. This method does not build a map, which we consider crucial for other downstream tasks.}{, although the proposed method does not build a map of the environment, which we consider crucial for downstream tasks, collaborative robotics and even path-planning, while the approach in \cite{loquercio2021learning} requires an operator to provide a reference trajectory to follow.}

A prominent direction in autonomous navigation are LiDAR-based systems, which have \replaced{a superior}{demonstrated improved} performance when compared with vision-based systems due to the unpaired depth perception capabilities of LiDAR sensors. A seminal work in this area was presented in \cite{tranzatto2022cerberus} \replaced{where they fuse visual and thermal imagery, LiDAR depth data, inertial cues, and kinematic data for state estimation based on \cite{khattak2020complementary}.}{by the winners of the DARPA Subterranean Challenge. To perform state estimation, they presented a multi-sensor fusion of visual and thermal imagery, LiDAR depth data, inertial cues, and kinematic pose estimates based on \cite{khattak2020complementary}.}

\replaced{Liu \textit{et al.} \cite{liu2022large} present a system that }{Another example is presented by Liu \textit{et al.} \cite{liu2022large},
which }loosely couples semantic LiDAR odometry and visual inertial states to estimate the odometry of the MAV.
They base their work on \cite{mohta2018experiments}, \cite{mohta2018fast} and they improve SLOAM~\cite{chen2020sloam} by incorporating visual-inertial odometry and a semantic segmentation network for LiDAR data. They also showcase their system in under-canopy forest navigation and \deleted{, to the best of our knowledge, it is the only system which} leverage SLAM instead of plain odometry\deleted{for state estimation}. Updating a planned path when odometry discontinuities occur upon loop-closure is a major issue for SLAM-based autonomous navigation systems. This is handled in \cite{liu2022large} by rigidly transforming the path based on the estimated drift upon loop-closure. This differs from our approach, where we elastically deform trajectory poses based on the estimated drift of the submaps the path traverses. This becomes important when planning through old submaps, as the drifts of individual submaps are not constant, but accumulate, and our elastic deformation adapts to the submaps' drift, keeping the trajectories safely in free space.

Compared to these systems, the main contribution of the proposed method is that it does not rely on LiDAR sensors but only on passive visual sensors, allowing a reduction of the system's mass, battery consumption and cost.
\begin{figure*}[t]
    \centering
    \rule{0pt}{0ex}
    \includegraphics[width=\textwidth, clip]{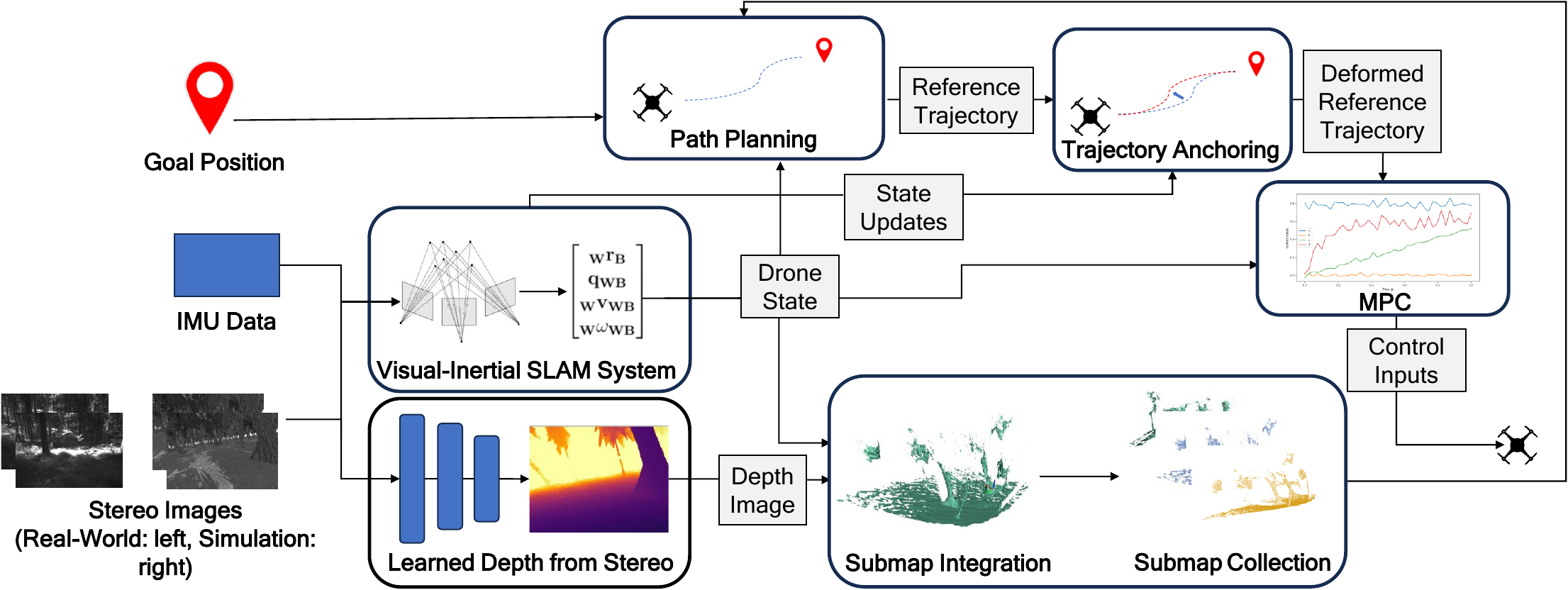}
    \caption{Overall approach overview.  Incoming stereo frames and IMU measurements are processed by the OKVIS2 VI-SLAM system. In parallel, the stereo images are processed by a CNN to output a depth map, which is then integrated into occupancy submaps. These in turn are used by a planner, generating a reference trajectory to be followed by the MPC. Upon state updates, especially pronounced after loop-closures, the trajectory is kept anchored to state estimates, i.e.\ moved and deformed in-line with the estimates and submaps.}
    \label{fig:architecture}
\end{figure*}

\section{Proposed approach}
\label{section:approach}

In this section we will present the different modules which build our system. Sections \ref{sec:MAV} -- \ref{sec:MPC} outline the hardware, notation, state estimation and control components of our system for completeness, while our proposed submapping interface and novel path planning and trajectory anchoring approach are described in Sections \ref{sec:SmInterface} -- \ref{subsec:anchoring}. A schematic overview is presented in Fig.~\ref{fig:architecture}


\subsection{MAV platform}
\label{sec:MAV}
We use the quadrotor shown in Fig.~\ref{fig:mesh_reconstruction}, consisting of a HolybroS500 V2 ARF-Kit airframe, an mRo PixRacer R15 flight controller, an Intel RealSense D455 camera (only using the IR stereo images and the IMU), and an NVIDIA Jetson Orin NX 16GB computer for on-board processing.

\subsection{Notation and Definitions}
\label{section:notation}
The used VI-SLAM \cite{leutenegger2022okvis2} tracks a moving body with a mounted IMU and several cameras relative to a static World coordinate frame $\cframe{W}$. The IMU coordinate frame is denoted as $\cframe{S}$, and the camera frames as $\cframe{C_i}$, with $i\in{\{1,2\}}$ as we use a stereo setup. Left-hand indices denote coordinate representation. Homogeneous position vectors (denoted in italics) can be
transformed with $\T{A}{B}$, meaning $\posh{A}{P} = \T{A}{B} \posh{B}{P}$. 

\subsection{State Estimation}
\label{subsec:state_estimation}

Our state estimation module is based on the existing VI-SLAM system OKVIS2. We will provide a brief recap of the key components of this work. For more details on OKVIS2, please refer to \cite{leutenegger2022okvis2}. OKVIS2 requires IMU measurements and stereo images as inputs to perform state estimation and build a sparse map of the environment, based on the visual keypoints extracted from the images. The state representation in this work is:
\begin{equation} \label{eqn}
\mbf{x}_{l} = \left[ \pos{W}{S}^{T}, \mbf{q}_{WS}^{T}, _{W}\mbf{v}^{T}, \mbf{b}_{g}^{T}, \mbf{b}_{a}^{T} \right] ^ {T},
\end{equation}
i.e.\ the position $\pos{W}{S}$ (of the IMU frame origin) expressed in World coordinates, its orientation quaternion $\mbf{q}_{WS}$, velocity $_{W}\mbf{v}$, and gyroscope and accelerometer biases $\mbf{b}_{g}$ and $\mbf{b}_{a}$, respectively.
A new state $\mbf{x}_{l}$ is estimated when a pair of stereo images is received.
In the estimator, two non-linear least squares optimizations are present: one executed in a real-time manner, a sliding window optimization runs over the last $T$ frames, $M$ keyframes, considering IMU measurements in-between, as well as reprojection errors and pose-graph errors formulated from marginalized co-observations; and a second optimization problem, using the exact same factor graph, is triggered on loop-closure in the background, which optimizes older states around the loop, too. This means that when processing a new frame, not only its state $\mbf x_l$ is estimated, but possibly many prior states $\mbf x_{l-k}$ are also updated.

\subsection{Model Predictive Controller}
\label{sec:MPC}
We use the linear model predictive controller (MPC) presented in \cite{tzoumanikasfully}, modified so it accepts the trajectory updates processed by the trajectory anchoring as described below in Sec.~\ref{subsec:anchoring}.  Previously, the system would receive the odometry states asynchronously from the trajectory states to be tracked. For correct trajectory tracking under the presence of major odometry changes, as it happens with loop-closures, the odometry update has to be processed synchronously with the deformed trajectory to ensure smooth and safe navigation. 

\subsection{Submap Interface}
\label{sec:SmInterface}
In order to allow safe path planning and navigation, we perform dense mapping using supereight2~\cite{funk2021multi}, an octree-based volumetric occupancy mapping framework.
It achieves efficient measurement integration and collision checking by leveraging the hierarchical structure of octrees.

As classical stereo-matching algorithms can fail in outdoor scenarios and only yield sparse output, we estimate depth from stereo images by fine-tuning \cite{xu2022attention} with synthetic outdoor data from \cite{tartanair2020iros}.
Given a set of depth images $\mbf{D}_t$ at a time step $t$, we generate an occupancy map. OKVIS2 computes a state $\mbf{x}_t$ from said stereo images, thus $\mbf{D}_t$ is associated with the pose estimates in $\mbf{x}_t$. Therefore we can use the pose of camera 1 in which $\mbf{D}_t$ is expressed, i.e. \ $\T{W}{C_{1,t}}=\T{W}{S_t}\T{S}{C_1}$, where $\T{S}{C_1}$ is the extrinsic calibration of camera 1. 

To account for inaccuracies and drift corrections from the SLAM system, it is essential to avoid monolithic volumetric maps. Otherwise, maps inconsistent with trajectory estimates are constructed, which are artificially cluttered with duplicated structure after corrections -- in essence, entirely useless for any form of safe navigation therein.
We adopt a submapping approach in which submaps are associated with keyframes from OKVIS2 and rigidly move with them.

\begin{figure}[t]
    \centering
    \rule{0pt}{1ex}
    \includegraphics[width=0.9\columnwidth]{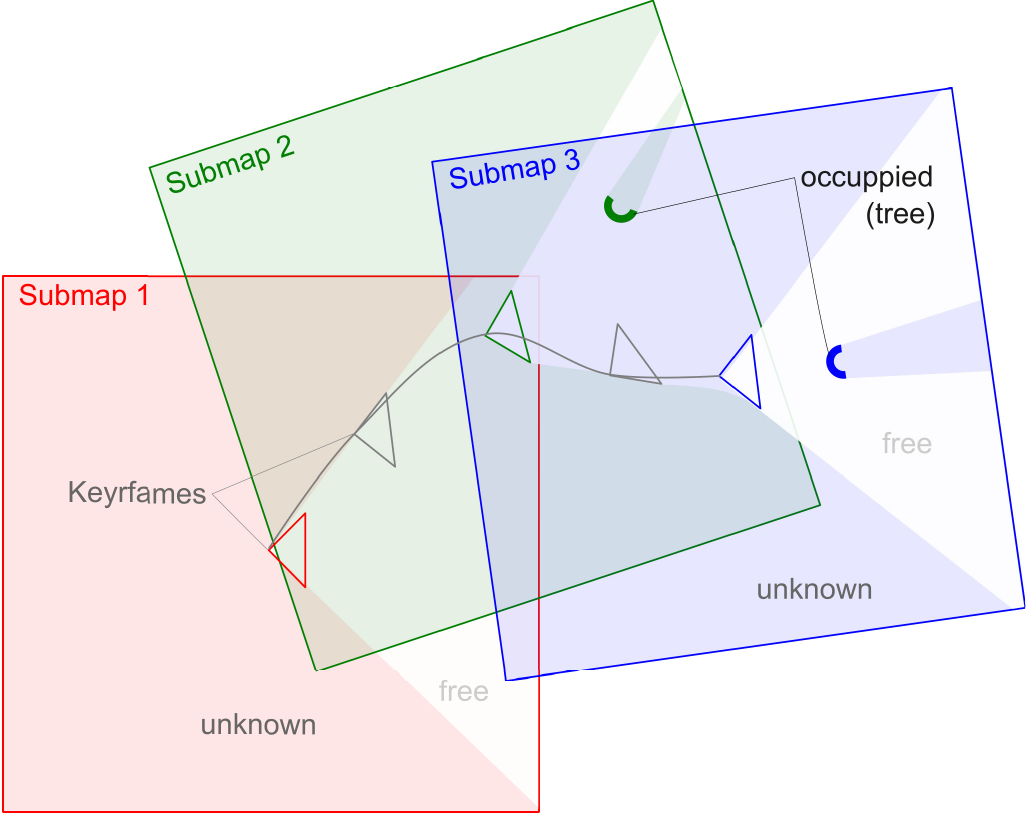}
    \caption{Submap generation example every two OKVIS2 keyframes.  Note the differentiation between unknown, free, and occupied space per submap. Submaps are aligned with the respective keyframes and move with them when OKVIS2 re-estimates them -- which is pronounced upon loop-closure.}
    \label{fig:submap_generation}
\end{figure}

The submap generation policy is based on the keyframe selection done by OKVIS2. Every $n$ keyframes, a new submap, with pose $\T{W}{S_k}$, is generated, where the upcoming depth images will be integrated as they are processed, with the $k$-th keyframe of the trajectory. An example of the submap generation policy can be found in Fig.~\ref{fig:submap_generation}. As both $\T{W}{S_k}$ and $\T{W}{C_{1,t}}$ are known, the relative transform $\T{S_k}{C_{1,t}}$ can be computed, i.e.\ the pose of the depth frame at time step $t$ relative to the submap coordinate frame corresponding to the pose at time step $k$, $\cframe{S_k}$, allowing depth frame integration.  
On keyframe state updates, the poses of submaps move accordingly.\looseness=-1

\subsection{Path Planning and Reference Trajectory Anchoring} 
\label{subsec:anchoring}


Our path planning is based on OMPL ~\cite{sucan2012open}, specifically using the Informed RRT*~\cite{informed_rrt_star} algorithm. To extend the path planner to the underlying submap structure, we assume that if a path segment (comprised of an equal-radius cylinder and a half-sphere at the end of the segment) is free in any of the maps from the subset of submaps $P$, it is a valid segment to navigate through. A segment is considered free if the log-odds occupancy value is negative, which only occurs in observed free-space. The submap subset $P$ is obtained via the covisibility graph of the state-estimator, that is, a graph with states as vertices that are connected by edges if they have covisible landmarks. We first compute the set $K$ of all keyframe states on the covisibility graph, that are connected by a path of at most 15 edges to the current state. $P$ then consists of the submaps that are anchored to one of the keyframe states in $K$, allowing us to reuse submaps after a loop-closure. \looseness=-1

Once a safe path is returned, it is converted into a trajectory, by accounting for the MAV's maximum velocity and acceleration, and undergoes our proposed trajectory anchoring, moving the reference in-line with updates from the state estimator. ``Jumps'' in state estimates would otherwise lead to erratic drone motion and ultimately to a crash, as the drone will aim to navigate to the previous trajectory state which represents a different environment location. Furthermore, we need to ensure that the planned reference trajectory remains continuous and in free space after the submap poses are updated. \looseness=-1

\begin{figure}[t]
    \centering
    \rule{0pt}{1ex}
    \includegraphics[width=0.86\columnwidth]{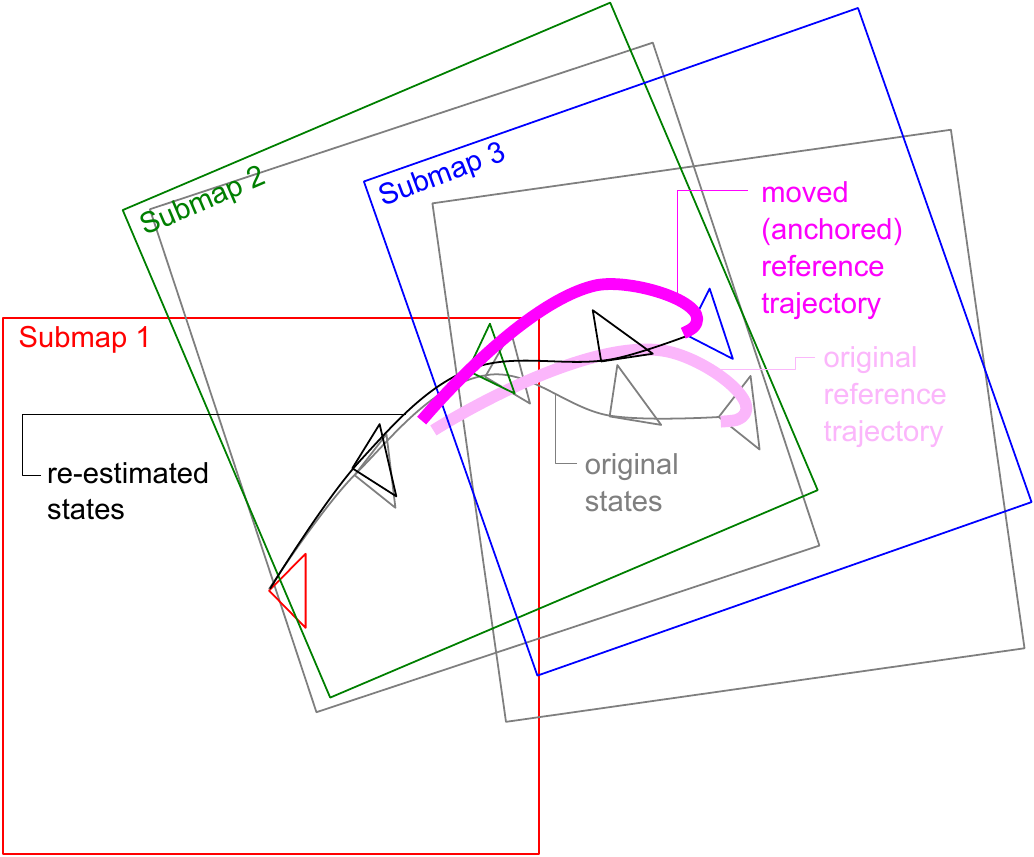}
    \caption{Reference trajectory anchoring.  When estimated states change, e.g.\ upon loop-closure, the planned reference trajectory moves in-line with submaps which are also anchored to keyframe states. The previous state trajectory, keyframe, and submap poses are drawn in gray, and the updated state trajectory in black. The previous trajectory (light pink) is moved and deformed (pink) in line with the submaps.}
    \label{fig:anchoring}
\end{figure}

Our approach, as visualized in Fig.~\ref{fig:anchoring}, anchors each reference state to a set $\mathcal{S}$ of states $\mbf{x}_s$, $s \in \{1, \ldots, S\}$ estimated by OKVIS2. The reference trajectory states will be updated as the anchor states $\mbf{x}_s$ update their corresponding poses.

A reference trajectory contains references (superscript ``ref'') $\mbf{x}^\text{ref}_j=({_W}\mbf{r}_{WS_j}^\text{ref}, \mbf{q}_{WS_j}^\text{ref}, {_W}\mbf{v}_j^\text{ref})$, with $j \in \{1, \ldots , J\}$ (and corresponding timestamps), which are anchored to the estimated OKVIS2 states in $\mathcal{S}$ via the algorithm of k-nearest neighbours, using the distance metric:
\begin{equation}
    d_{j,s} = ||{_W}\mbf{r}_{WS_j}^\text{ref} - \pos{W}{S_s}||.
\end{equation}
We associate every pair $j,s$ with the relative transformation $\T{S_s}{S_j}$ and a weight $w_{j,s}$:
\begin{equation} \label{weight_equation}
w_{j,s} = \frac{{1}/{d_{j,s}}}{\sum_{j=1}^{S}{1}/{d_{j,s}}}.
\end{equation}
On state update, trajectory state positions are updated with:\looseness=-1
\begin{equation}
    {_W}\mbf{r}_{WS_j}^\text{ref'} = \sum_{s = 1}^{S} w_{j,s} \hspace{0.2em} \T{W}{S_s} \hspace{0.2em} \pos{S_s}{S_j},
\end{equation}
where $\T{W}{S_s}$ is the latest anchor pose as estimated by OKVIS2.
The orientation $\mbf{q}_{WS_j}^\text{ref'}$ is updated to the weighted average of $\mbf{q}_{S_sS_j}, \, s \in \{1 \dots S\}$ using the previously computed weights by following the method from \cite{markley2007averaging}. We also apply the respective orientation changes to the velocities.

\section{Evaluation}
\label{section:eval}

We quantitatively validated our method in several simulation experiments and performed a qualitative evaluation in a real-world forest. We used the same parameters for both simulated and real-world experiments: a 10 cm map resolution, 0.5 s of planning time, a 3 m/s maximum MAV velocity and identical OVKIS2 parameters. Loop-closures were present in all cases, necessitating trajectory anchoring.

\begin{figure}
    \centering
    \includegraphics[width=\columnwidth]{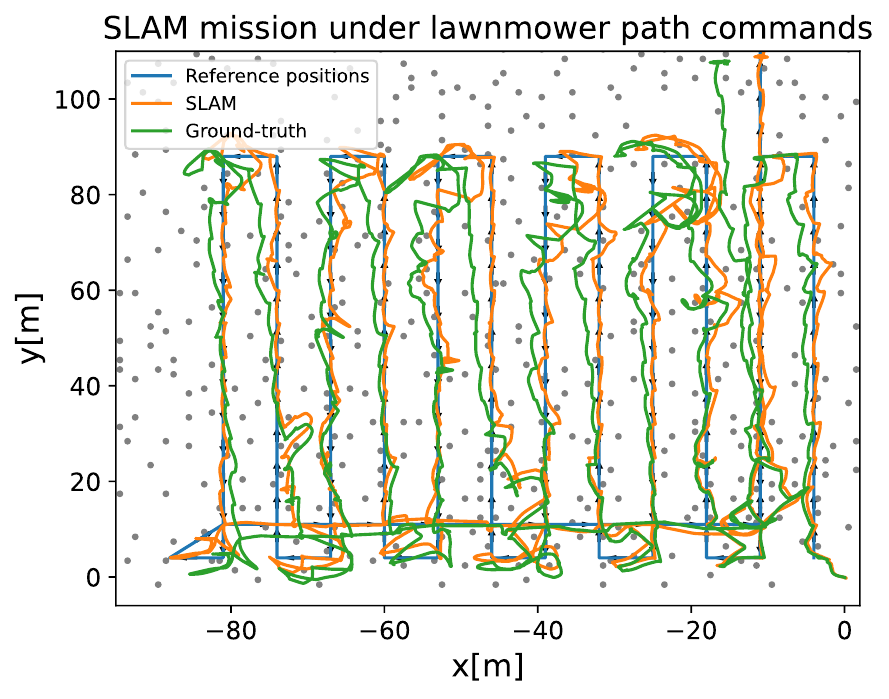}
    \includegraphics[width=\columnwidth]{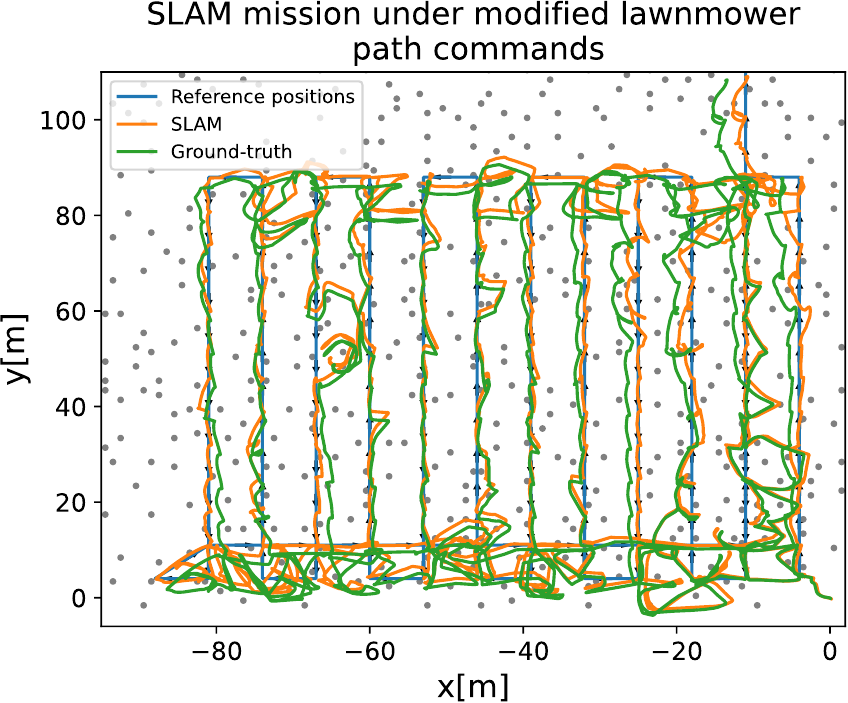}
    \caption{Overhead views of two simulation experiments, following a lawnmower pattern (top) and using the proposed modified lawnmower pattern (bottom) which triggers more loop-closures.
    The images show the MAV reference positions, the live trajectory estimated by OKVIS2 and the corresponding ground-truth trajectory.
    The reference positions can not always be accurately tracked due to trees (gray dots).
    }
    \label{fig:mission}
\end{figure}

\subsection{Simulation Experiments}
We use Ignition Gazebo as our simulation environment and we leverage PX4's simulation in the loop capabilities to interact with the drone, equivalently to the real-world setup. In this simulation, we have created an artificial forest that contains 620 trees in a square-shaped surface area of 16'384 m$^2$, i.e. 378 trees per hectare. According to \cite{swiss_forest}, Swiss forests have an average tree density of 690 trees for the same surface area, validating the realism of our simulation.

To demonstrate the capability of our system to fly long missions in the presence of loop-closures, and the superior quality of the maps generated by a system leveraging SLAM rather than just VIO, we evaluate on two mission types: one using a classic lawnmower pattern reference path and the other a slight modification of the former which triggers more loop-closures by revisiting previously mapped areas.
The presence of large loop-closures showcases the need for the trajectory anchoring module to ensure safe navigation.
In total, 12 missions were executed, \emph{without a single collision}.
The lawnmower path and its modified version had an average length of 1869.78 m and 2363.64 m  respectively, and trajectory examples from these missions are shown in Fig.~\ref{fig:mission}.
The mean error of the depth network is of 0.63 m for estimates under 6.5 m, the maximum depth for a submap update.

\begin{figure}[t]
    \centering
    \includegraphics[width=\columnwidth]{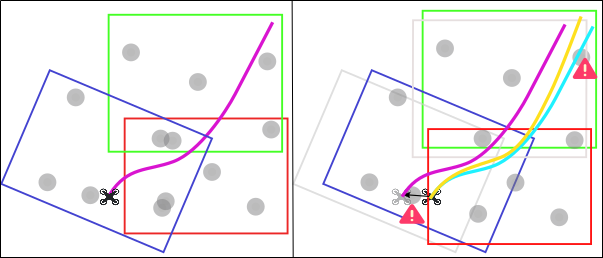}
    \caption{
    Planned trajectory before loop-closure (left) and trajectory adaptation strategies after loop-closure (right), with obstacles as gray disks. The adaptation strategies are: no adaptation (purple), rigid transformation (cyan) and the proposed trajectory anchoring (yellow).
    Without adaptation, the MAV flies towards its estimated pose before loop-closure (black arrow), which is unsafe. Rigidly transforming the trajectory does not account for the accumulated drift of each submap along it, leading to unsafe trajectories.
    }
    \label{fig:trajectory_anchoring_simu}
\end{figure}



To showcase the superior mapping quality when leveraging SLAM, we evaluate the accuracy and completeness of the meshes obtained via SLAM, VIO and the best-case-scenario of using both ground-truth poses and depth maps, against the ground-truth mesh.
The accuracy measured by the root-mean-square error (RMSE) and completeness by the percentage of the ground-truth mesh vertices for which a reconstructed mesh polygon is within 20 or 50 cm.
The absolute trajectory error is computed between the estimated and ground-truth trajectories.


\begin{figure}[]
    \centering
    \includegraphics[width=\columnwidth]{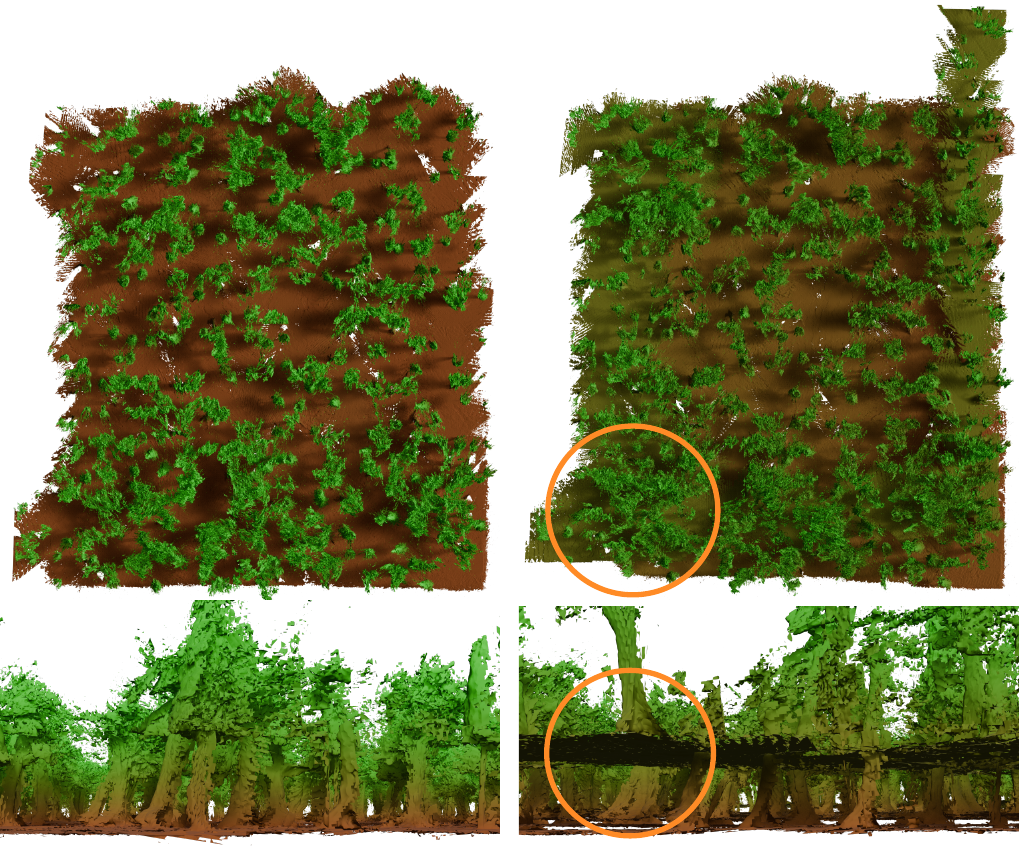}
    \caption{
    \added{Top and side view, from the same perspective, of the final meshes after a modified lawnmower mission. The left and right columns show the reconstructions obtained with SLAM and plain VIO respectively. The orange circles show areas where drift has a clear effect on the reconstruction quality, making it unusable for any downstream task.} 
    }
    \label{fig:qualitative}
\end{figure}

The quantitative evaluation results are shown in Table \ref{tab:mesh_accuracy}. The best-case-scenario mesh reconstructions do not achieve 100\% completeness as the environment is not fully observable. Drift correction due to loop-closures improves both the reconstructed mesh quality and the trajectory accuracy in both mission types, with a larger improvement using the modified lawnmower pattern, due to the higher amount of homogeneously triggered loop-closures. \added{A qualitative comparison of the reconstructed meshes via SLAM and VIO is presented in Fig. \ref{fig:qualitative}}. We benchmark against the state-of-the-art, submap-based, dense mapping framework \voxgraph~\cite{reijgwart2019voxgraph}, which performs submap to submap alignment, and is the mapping framework used in the autonomous exploration system presented in \cite{glocal}. \added{We do not benchmark against \cite{real} since this method is not suited to such large environments as it does not handle trajectory deformations upon loop closure. Furthermore, it requires a monolithic map that would be degraded by drift, and the maximum flight speeds of 0.5 m/s would make some of the missions last for more than 1 hour, in the best-case scenario}. We compare at a 20 cm map resolution, since \voxgraph requires more than the available 45 GB of RAM for mapping at 10 cm. At this resolution, our system requires 18.4 GB and 25.3 GB on average for the lawnmower and modified lawnmower missions respectively, showcasing a superior capability to scale to large environments.
This is partly due to \voxgraph's time-based submap generation policy that leads to more submaps than our visual overlap criterion. 
To evaluate at 20 cm resolution we use the data from the missions mapped at 10 cm, since collisions occurred against unmapped thin structures when using coarse maps. Our approach produces more complete maps in both missions and has substantially higher accuracy in the modified lawnmower mission. When following the lawnmower pattern, \voxgraph had a marginally more accurate map as it did not map thin structures, while our framework noisily represented them, obtaining more complete maps at the expense of accuracy.

\begin{table}[t]
\caption{Percentage of completed missions when performing navigation through prior submaps upon loop-closure.
}
\centering
\begin{tabular}{ |c|c| } 
\hline
Deformation strategy & Completed missions (\%) $\uparrow$ \\
\hline
 No deformation & 40 \\ 
 Rigid deformation & 75 \\ 
 Trajectory anchoring (ours) & 95 \\ 
 \hline
\end{tabular}
\label{tab:traj_anchoring_ablation}
\end{table}

\begin{table}[t]
\caption{Mean and standard deviation of mesh accuracy, mean mesh completeness and absolute trajectory error (ATE) for all missions using \voxgraph, SLAM, VIO and ground-truth for both poses and depth images (GT SLAM/VIO).}
\centering
\setlength\tabcolsep{4pt}
\rule{0pt}{0ex}
\resizebox{!}{56pt}{ 
\begin{tabular}{lc|c|cc|c|}
\cline{3-6}
& & \multirow{2}{*}{RMSE (m) $\downarrow$} & \multicolumn{2}{c|}{Compl. (\%) $\uparrow$} & \multirow{2}{*}{ATE (m) $\downarrow$} \\ \cline{4-5}
& & & \multicolumn{1}{c|}{20 cm} & 50 cm & \\ \hline
\multicolumn{1}{|l|}{\voxgraph\cite{reijgwart2019voxgraph} (20cm)} & \multirow{6}{*}{Lawnmower} & 0.192$\pm$0.083 & \multicolumn{1}{l|}{15.06} & 26.69 & 1.30$\pm$0.12 \\
\multicolumn{1}{|l|}{SLAM (20cm)} & & 0.217$\pm$0.050 & \multicolumn{1}{l|}{19.73} & 36.38 & 0.80$\pm$0.07 \\
\multicolumn{1}{|l|}{SLAM (10cm)} & & 0.178$\pm$0.046 & \multicolumn{1}{l|}{25.55} & 46.59 & 0.74$\pm$0.30 \\
\multicolumn{1}{|l|}{VIO (10cm)} & & 0.284$\pm$0.024 & \multicolumn{1}{c|}{24.23}    & 44.28    & 1.33$\pm$0.17                    \\
\multicolumn{1}{|l|}{\dimt{GT SLAM (10cm)}} &                                     & \dimt{0.058$\pm$0.086}                                      & \multicolumn{1}{c|}{\dimt{34.60}}    & \dimt{51.97}    & \dimt{-}                    \\
\multicolumn{1}{|l|}{\dimt{GT VIO (10cm)}}  &                                     & \dimt{0.059$\pm$0.090}                                      & \multicolumn{1}{c|}{\dimt{34.26}}    & \dimt{51.57}    & \dimt{-}                    \\ \hline
\multicolumn{1}{|l|}{\voxgraph\cite{reijgwart2019voxgraph} (20cm)} & \multirow{6}{*}{\begin{tabular}[c]{@{}c@{}}Modified \\ Lawnmower\end{tabular}} & 0.265$\pm$0.048 & \multicolumn{1}{l|}{14.63} & 27.11 & 0.92$\pm$0.04 \\
\multicolumn{1}{|l|}{SLAM (20cm)} & & 0.195$\pm$0.061 & \multicolumn{1}{l|}{24.21} & 40.08 & 0.43$\pm$0.09 \\
\multicolumn{1}{|l|}{SLAM (10cm)}    & & 0.118$\pm$0.003 & \multicolumn{1}{c|}{30.91} & 50.13 & 0.51$\pm$0.07\\
\multicolumn{1}{|l|}{VIO (10cm)}     &  & 0.334$\pm$0.096 & \multicolumn{1}{c|}{24.90}    & 46.38    & 1.26$\pm$0.54                     \\
\multicolumn{1}{|l|}{\dimt{GT SLAM (10cm)}} &                                     & \dimt{0.058$\pm$0.082}                                      & \multicolumn{1}{c|}{\dimt{36.71}}    & \dimt{54.65}    & \dimt{-}                    \\
\multicolumn{1}{|l|}{\dimt{GT VIO (10cm)}}  &                                     & \dimt{0.058$\pm$0.082}                                      & \multicolumn{1}{c|}{\dimt{36.36}}    & \dimt{54.28}    & \dimt{-}                    \\ \hline
\end{tabular}
}
\label{tab:mesh_accuracy}
\end{table}

In another set of experiments, we demonstrate the benefits of trajectory anchoring upon loop-closure, especially when reusing old submaps for navigation. We commanded the drone to fly 80 m forward, building a collection of submaps, return to its starting position and then fly 80 m forward again, triggering loop-closures. We added Gaussian white noise to the IMU, while leaving it unaccounted for in the state-estimator, to increase the otherwise negligible drift for this trajectory length. We evaluated three approaches to trajectory adaptation upon state updates: no adaptation, \added{as done in \cite{real}}; rigid transformation based on the estimated drift upon loop-closure, as proposed by \cite{liu2022large}; and our proposed trajectory anchoring. The adaptation strategies are visualized in Fig \ref{fig:trajectory_anchoring_simu}. The experiment was performed 20 times for each adaptation strategy and our approach achieved the highest number of missions without collisions as shown in Table \ref{tab:traj_anchoring_ablation}. The only collision with trajectory anchoring was due to the state-estimator accumulating high drift. Rigid deformation is suitable for LiDAR SLAM where drift can be assumed constant in a local neighborhood, unlike vision-based SLAM.

\begin{figure}
     \centering
     \includegraphics[width=\columnwidth, trim={0.6cm, 0.0cm, 0.0cm, 0.55cm}, clip]{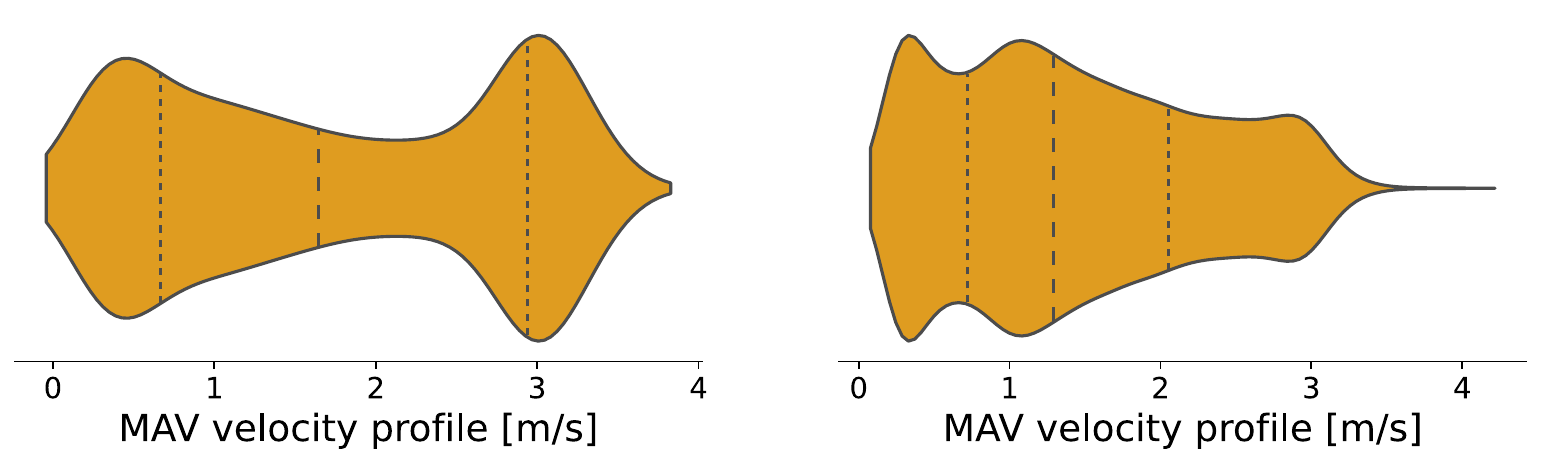}
     \caption{
     Velocity profile for a trajectory anchoring benchmark mission (left) and a modified lawnmower mission (right).
     }
     
     \label{fig:statistics_simulation}
 \end{figure}

Fig. ~\ref{fig:statistics_simulation} shows the ground-truth velocity profiles of two missions: the modified lawnmower pattern and the mission used for trajectory anchoring evaluation. The differences in the velocity profiles are due to the usage of prior submaps for navigation. In the lawnmower missions, the waypoint was towards areas which are unmapped, producing shorter planned trajectories, while in the missions where we benchmarked trajectory anchoring, longer paths were executed, as waypoints were in previously mapped areas, reaching the peak velocity for longer periods of time.

\renewcommand{\arraystretch}{1}

\subsection{Real World Experiments}
\label{subsec:real_world_experiments}

We performed real-world experiments in a forest with an approximate density of 467 trees per hectare, denser than the simulated one.
The MAV goal poses were manually set by an operator.
The overall mission consisted of doing a path forward, then turning around and returning in the opposite direction as shown in Fig.~\ref{fig:mesh_reconstruction}. During the return phase, the planned paths were longer, because prior submaps could be used for path planning. Near home, an extra detour was done to change the orientation to the initial one and trigger a final, long loop-closure. OKVIS2 processed stereo images at 15 Hz and depth images were generated at 5 Hz on average. A more detailed runtime analysis of the different system components can be found in Table \ref{tab:process-timings}. As it can be seen, the depth inference is the bottleneck due to the limited compute power of the onboard GPU. Nonetheless, due to the multi-threaded nature of the system, the state-estimator is decoupled from the depth inference. This allows the state-estimator to process all the incoming stereo images even though the corresponding depth images can not be obtained.

\setlength{\tabcolsep}{4.0pt}
\begin{table}[h]
\caption{Average processing time per-image.}
\label{tab:process-timings}
\centering
\renewcommand{\arraystretch}{1} 
\begin{tabular}{c|c}
Stage & Average time {[}ms{]}  \\ \hline
Depth inference        & 217 \\
\hline
Depth integration & 56 \\
Trajectory anchoring &  0.19 \\ 
Submap pose updates & 0.002 \\  \hline
Keypoint detection & 13 \\
Landmark matching & 39 \\ \hline
Graph optimization & 52 \\
\end{tabular}

\begin{tablenotes}
   \item Horizontal lines separate processing steps that run in parallel threads.
\end{tablenotes}
\end{table}

We have roughly determined the drift between the start and end positions to be less than 1 m \emph{before} loop-closure -- since the drone was commanded back to the start. According to the OKVIS2 odometry, the total distance travelled during the mission was 226.71 m. Thus, the estimator position drift amounts to $<0.5\%$ of the distance travelled.

The maximum velocity achieved was 4 m/s according to the OKVIS2 odometry, although the average velocity was 1.2 m/s with the lower and upper quartiles being 0.6 m/s and 1.8 m/s, respectively. 
\emph{No collisions} occurred and \emph{no interventions} had to be made.

\section{Conclusion}
\label{section:conclusion}

In this paper, we present an MAV system which relies solely on passive visual sensors and an IMU to perform autonomous under-canopy navigation in forests. Onboard visual-inertial SLAM is in charge of MAV state estimation. We propose a volumetric occupancy submapping system to achieve a scalable mapping representation of the environment and reduce the reconstruction errors which arise in monolithic maps under drift and brusquely changing state estimates. We also introduce a trajectory anchoring approach to ensure safe navigation upon abrupt changes in odometry due to loop-closures. To demonstrate the method, we validated the system both in simulation and in a real-world environment -- demonstrating safe flight at up to 3 m/s without a single collision or unsafe planning instance.

\IEEEtriggeratref{24}
\bibliographystyle{IEEEtran}
\bibliography{references}

\begin{thebibliography}{10}
\providecommand{\url}[1]{#1}
\csname url@samestyle\endcsname
\providecommand{\newblock}{\relax}
\providecommand{\bibinfo}[2]{#2}
\providecommand{\BIBentrySTDinterwordspacing}{\spaceskip=0pt\relax}
\providecommand{\BIBentryALTinterwordstretchfactor}{4}
\providecommand{\BIBentryALTinterwordspacing}{\spaceskip=\fontdimen2\font plus
\BIBentryALTinterwordstretchfactor\fontdimen3\font minus
  \fontdimen4\font\relax}
\providecommand{\BIBforeignlanguage}[2]{{%
\expandafter\ifx\csname l@#1\endcsname\relax
\typeout{** WARNING: IEEEtran.bst: No hyphenation pattern has been}%
\typeout{** loaded for the language `#1'. Using the pattern for}%
\typeout{** the default language instead.}%
\else
\language=\csname l@#1\endcsname
\fi
#2}}
\providecommand{\BIBdecl}{\relax}
\BIBdecl

\bibitem{liu2022large}
X.~Liu, G.~V. Nardari, F.~C. Ojeda, Y.~Tao, A.~Zhou, T.~Donnelly, C.~Qu, S.~W.
  Chen, R.~A. Romero, C.~J. Taylor \emph{et~al.}, ``Large-scale autonomous
  flight with real-time semantic {SLAM} under dense forest canopy,'' \emph{IEEE
  Robotics and Automation Letters}, vol.~7, no.~2, pp. 5512--5519, 2022.

\bibitem{tranzatto2022cerberus}
M.~Tranzatto, T.~Miki, M.~Dharmadhikari, L.~Bernreiter, M.~Kulkarni,
  F.~Mascarich, O.~Andersson, S.~Khattak, M.~Hutter, R.~Siegwart \emph{et~al.},
  ``{CERBERUS} in the {DARPA} {S}ubterranean {C}hallenge,'' \emph{Science
  Robotics}, vol.~7, no.~66, p. eabp9742, 2022.

\bibitem{faessler2016autonomous}
M.~Faessler, F.~Fontana, C.~Forster, E.~Mueggler, M.~Pizzoli, and
  D.~Scaramuzza, ``Autonomous, vision-based flight and live dense {3D} mapping
  with a quadrotor micro aerial vehicle,'' \emph{Journal of Field Robotics},
  vol.~33, no.~4, pp. 431--450, 2016.

\bibitem{oleynikova2020open}
H.~Oleynikova, C.~Lanegger, Z.~Taylor, M.~Pantic, A.~Millane, R.~Siegwart, and
  J.~Nieto, ``An open-source system for vision-based micro-aerial vehicle
  mapping, planning, and flight in cluttered environments,'' \emph{Journal of
  Field Robotics}, vol.~37, no.~4, pp. 642--666, 2020.

\bibitem{schneider2018maplab}
T.~Schneider, M.~Dymczyk, M.~Fehr, K.~Egger, S.~Lynen, I.~Gilitschenski, and
  R.~Siegwart, ``maplab: {A}n open framework for research in visual-inertial
  mapping and localization,'' \emph{IEEE Robotics and Automation Letters},
  vol.~3, no.~3, pp. 1418--1425, 2018.

\bibitem{voxblox}
H.~Oleynikova, Z.~Taylor, M.~Fehr, R.~Siegwart, and J.~Nieto, ``Voxblox:
  {I}ncremental {3D} {E}uclidean signed distance fields for on-board {MAV}
  planning,'' in \emph{2017 IEEE/RSJ International Conference on Intelligent
  Robots and Systems (IROS)}, 2017, pp. 1366--1373.

\bibitem{millane2018cblox}
A.~Millane, Z.~Taylor, H.~Oleynikova, J.~Nieto, R.~Siegwart, and C.~Cadena,
  ``C-blox: {A} scalable and consistent {TSDF}-based dense mapping approach,''
  in \emph{2018 IEEE/RSJ International Conference on Intelligent Robots and
  Systems (IROS)}, 2018.

\bibitem{lin2018autonomous}
Y.~Lin, F.~Gao, T.~Qin, W.~Gao, T.~Liu, W.~Wu, Z.~Yang, and S.~Shen,
  ``Autonomous aerial navigation using monocular visual-inertial fusion,''
  \emph{Journal of Field Robotics}, vol.~35, no.~1, pp. 23--51, 2018.

\bibitem{hornung2013octomap}
A.~Hornung, K.~M. Wurm, M.~Bennewitz, C.~Stachniss, and W.~Burgard,
  ``{OctoMap}: {An} efficient probabilistic {3D} mapping framework based on
  octrees,'' \emph{Autonomous robots}, vol.~34, pp. 189--206, 2013.

\bibitem{Cieslewski2017}
T.~Cieslewski, E.~Kaufmann, and D.~Scaramuzza, ``Rapid exploration with
  multi-rotors: {A} frontier selection method for high speed flight,'' in
  \emph{2017 IEEE/RSJ International Conference on Intelligent Robots and
  Systems (IROS)}, 2017, pp. 2135--2142.

\bibitem{Zichao2018}
Z.~Zhang and D.~Scaramuzza, ``Perception-aware receding horizon navigation for
  {MAV}s,'' in \emph{2018 IEEE International Conference on Robotics and
  Automation (ICRA)}, 2018, pp. 2534--2541.

\bibitem{foehn2022agilicious}
P.~Foehn, E.~Kaufmann, A.~Romero, R.~Penicka, S.~Sun, L.~Bauersfeld,
  T.~Laengle, G.~Cioffi, Y.~Song, A.~Loquercio \emph{et~al.}, ``Agilicious:
  Open-source and open-hardware agile quadrotor for vision-based flight,''
  \emph{Science robotics}, vol.~7, no.~67, p. eabl6259, 2022.

\bibitem{campos2021autonomous}
L.~Campos-Mac{\'\i}as, R.~Aldana-L{\'o}pez, R.~de~la Guardia, J.~I.
  Parra-Vilchis, and D.~G{\'o}mez-Guti{\'e}rrez, ``Autonomous navigation of
  mavs in unknown cluttered environments,'' \emph{Journal of Field Robotics},
  vol.~38, no.~2, pp. 307--326, 2021.

\bibitem{alarcon2021efficient}
E.~P.~H. Alarc{\'o}n, D.~B. Ghavifekr, G.~Baris, M.~Mugnai, M.~Satler, and
  C.~A. Avizzano, ``An efficient object-oriented exploration algorithm for
  unmanned aerial vehicles,'' in \emph{2021 International Conference on
  Unmanned Aircraft Systems (ICUAS)}.\hskip 1em plus 0.5em minus 0.4em\relax
  IEEE, 2021, pp. 330--337.

\bibitem{reijgwart2024waverider}
V.~Reijgwart, M.~Pantic, R.~Siegwart, and L.~Ott, ``Waverider: Leveraging
  hierarchical, multi-resolution maps for efficient and reactive obstacle
  avoidance,'' in \emph{2024 IEEE International Conference on Robotics and
  Automation (ICRA)}.\hskip 1em plus 0.5em minus 0.4em\relax IEEE, 2024, pp.
  13\,157--13\,163.

\bibitem{bircher2016receding}
A.~Bircher, M.~Kamel, K.~Alexis, H.~Oleynikova, and R.~Siegwart, ``Receding
  horizon" next-best-view" planner for 3d exploration,'' in \emph{2016 IEEE
  international conference on robotics and automation (ICRA)}.\hskip 1em plus
  0.5em minus 0.4em\relax IEEE, 2016, pp. 1462--1468.

\bibitem{liu2024omninxt}
P.~Liu, C.~Feng, Y.~Xu, Y.~Ning, H.~Xu, and S.~Shen, ``Omninxt: A fully
  open-source and compact aerial robot with omnidirectional visual
  perception,'' in \emph{2024 IEEE/RSJ International Conference on Intelligent
  Robots and Systems (IROS)}.\hskip 1em plus 0.5em minus 0.4em\relax IEEE,
  2024, pp. 10\,605--10\,612.

\bibitem{real}
E.~M. Lee, J.~Choi, H.~Lim, and H.~Myung, ``Real: Rapid exploration with active
  loop-closing toward large-scale 3d mapping using uavs,'' in \emph{2021
  IEEE/RSJ International Conference on Intelligent Robots and Systems (IROS)},
  2021, pp. 4194--4198.

\bibitem{reijgwart2019voxgraph}
V.~Reijgwart, A.~Millane, H.~Oleynikova, R.~Siegwart, C.~Cadena, and J.~Nieto,
  ``Voxgraph: Globally consistent, volumetric mapping using signed distance
  function submaps,'' \emph{IEEE Robotics and Automation Letters}, vol.~5,
  no.~1, pp. 227--234, 2019.

\bibitem{glocal}
L.~Schmid, V.~Reijgwart, L.~Ott, J.~Nieto, R.~Siegwart, and C.~Cadena, ``A
  unified approach for autonomous volumetric exploration of large scale
  environments under severe odometry drift,'' \emph{IEEE Robotics and
  Automation Letters}, vol.~6, no.~3, pp. 4504--4511, 2021.

\bibitem{zhou2021ego}
X.~Zhou, J.~Zhu, H.~Zhou, C.~Xu, and F.~Gao, ``Ego-swarm: {A} fully autonomous
  and decentralized quadrotor swarm system in cluttered environments,'' in
  \emph{2021 IEEE international conference on robotics and automation
  (ICRA)}.\hskip 1em plus 0.5em minus 0.4em\relax IEEE, 2021, pp. 4101--4107.

\bibitem{loquercio2021learning}
A.~Loquercio, E.~Kaufmann, R.~Ranftl, M.~M{\"u}ller, V.~Koltun, and
  D.~Scaramuzza, ``Learning high-speed flight in the wild,'' \emph{Science
  Robotics}, vol.~6, no.~59, p. eabg5810, 2021.

\bibitem{zhou2020ego}
X.~Zhou, Z.~Wang, H.~Ye, C.~Xu, and F.~Gao, ``Ego-planner: {A}n {ESDF}-free
  gradient-based local planner for quadrotors,'' \emph{IEEE Robotics and
  Automation Letters}, vol.~6, no.~2, pp. 478--485, 2020.

\bibitem{khattak2020complementary}
S.~Khattak, H.~Nguyen, F.~Mascarich, T.~Dang, and K.~Alexis, ``Complementary
  multi--modal sensor fusion for resilient robot pose estimation in
  subterranean environments,'' in \emph{2020 International Conference on
  Unmanned Aircraft Systems (ICUAS)}.\hskip 1em plus 0.5em minus 0.4em\relax
  IEEE, 2020, pp. 1024--1029.

\bibitem{mohta2018experiments}
K.~Mohta, K.~Sun, S.~Liu, M.~Watterson, B.~Pfrommer, J.~Svacha, Y.~Mulgaonkar,
  C.~J. Taylor, and V.~Kumar, ``Experiments in fast, autonomous, {GPS}-denied
  quadrotor flight,'' in \emph{2018 IEEE International Conference on Robotics
  and Automation (ICRA)}.\hskip 1em plus 0.5em minus 0.4em\relax IEEE, 2018,
  pp. 7832--7839.

\bibitem{mohta2018fast}
K.~Mohta, M.~Watterson, Y.~Mulgaonkar, S.~Liu, C.~Qu, A.~Makineni, K.~Saulnier,
  K.~Sun, A.~Zhu, J.~Delmerico \emph{et~al.}, ``Fast, autonomous flight in
  {GPS}-denied and cluttered environments,'' \emph{Journal of Field Robotics},
  vol.~35, no.~1, pp. 101--120, 2018.

\bibitem{chen2020sloam}
S.~W. Chen, G.~V. Nardari, E.~S. Lee, C.~Qu, X.~Liu, R.~A.~F. Romero, and
  V.~Kumar, ``Sloam: Semantic lidar odometry and mapping for forest
  inventory,'' \emph{IEEE Robotics and Automation Letters}, vol.~5, no.~2, pp.
  612--619, 2020.

\bibitem{leutenegger2022okvis2}
S.~Leutenegger, ``{OKVIS2}: {R}ealtime scalable visual-inertial {SLAM} with
  loop closure,'' \emph{arXiv preprint arXiv:2202.09199}, 2022.

\bibitem{tzoumanikasfully}
D.~Tzoumanikas, W.~Li, M.~Grimm, K.~Zhang, M.~Kovac, and S.~Leutenegger,
  ``Fully autonomous micro air vehicle flight and landing on a moving target
  using visual--inertial estimation and model-predictive control,''
  \emph{Journal of Field Robotics}, vol.~36, no.~1, pp. 49--77, 2019.

\bibitem{funk2021multi}
N.~Funk, J.~Tarrio, S.~Papatheodorou, M.~Popovi{\'c}, P.~F. Alcantarilla, and
  S.~Leutenegger, ``Multi-resolution {3D} mapping with explicit free space
  representation for fast and accurate mobile robot motion planning,''
  \emph{IEEE Robotics and Automation Letters}, vol.~6, no.~2, pp. 3553--3560,
  2021.

\bibitem{xu2022attention}
G.~Xu, J.~Cheng, P.~Guo, and X.~Yang, ``Attention concatenation volume for
  accurate and efficient stereo matching,'' in \emph{Proceedings of the
  IEEE/CVF Conference on Computer Vision and Pattern Recognition}, 2022, pp.
  12\,981--12\,990.

\bibitem{tartanair2020iros}
W.~Wang, D.~Zhu, X.~Wang, Y.~Hu, Y.~Qiu, C.~Wang, Y.~Hu, A.~Kapoor, and
  S.~Scherer, ``{TartanAir}: {A} dataset to push the limits of visual {SLAM},''
  in \emph{2020 IEEE/RSJ International Conference on Intelligent Robots and
  Systems (IROS)}, 2020.

\bibitem{sucan2012open}
I.~A. Sucan, M.~Moll, and L.~E. Kavraki, ``The open motion planning library,''
  \emph{IEEE Robotics \& Automation Magazine}, vol.~19, no.~4, pp. 72--82,
  2012.

\bibitem{informed_rrt_star}
J.~D. Gammell, S.~S. Srinivasa, and T.~D. Barfoot, ``Informed {RRT*}: {O}ptimal
  sampling-based path planning focused via direct sampling of an admissible
  ellipsoidal heuristic,'' in \emph{IEEE/RSJ International Conference on
  Intelligent Robots and Systems}, Chicago, IL, USA, September 2014, pp.
  2997--3004.

\bibitem{markley2007averaging}
F.~L. Markley, Y.~Cheng, J.~L. Crassidis, and Y.~Oshman, ``Averaging
  quaternions,'' \emph{Journal of Guidance, Control, and Dynamics}, vol.~30,
  no.~4, pp. 1193--1197, 2007.

\bibitem{swiss_forest}
{Swiss National Forest Inventory},
  \url{https://lfi.ch/publikationen/publ/LFI_Flyer-en.pdf}, 2021.

\end{thebibliography}
\end{document}